\pdfoutput=1
\documentclass[10pt,twocolumn,letterpaper]{article}

\usepackage{cvpr}
\usepackage{times}
\usepackage{epsfig}
\usepackage{graphicx}
\usepackage{amsmath}
\usepackage{booktabs}
\usepackage{tabularx}
\usepackage{multirow}
\usepackage{multicol}
\usepackage{amssymb}
\usepackage[normalem]{ulem}
\usepackage{xcolor}

\usepackage[breaklinks=true,bookmarks=false]{hyperref}
\usepackage{authblk}

\cvprfinalcopy %

\usepackage{mathtools}
\newcommand\Mycomb[2][n]{\prescript{#1\mkern-0.8mu}{}C_{#2}}

\newcommand{\vqarep}[0]{VQA-Rephrasings\xspace}

\newcommand{\meetnew}[1]{\textcolor{black}{#1}}
\newcommand{\meetnewn}[1]{\textcolor{black}{#1}}

\newcommand{\todo}[1]{\textcolor{green}{}}

\def\cvprPaperID{3454} %

\newenvironment{packed_itemize}{
\begin{list}{\labelitemi}{\leftmargin=2em}
\vspace{-6pt}
 \setlength{\itemsep}{0pt}
 \setlength{\parskip}{0pt}
 \setlength{\parsep}{0pt}
}{\end{list}}

\begin{document}

\title{Cycle-Consistency for Robust Visual Question Answering}

\author{Meet Shah\textsuperscript{1}
}
\author{Xinlei Chen\textsuperscript{1}}
\author{Marcus Rohrbach\textsuperscript{1}}
\author{Devi Parikh\textsuperscript{1,2}}

\affil{\textsuperscript{1}Facebook AI Research, \textsuperscript{2}Georgia Institute of Technology
\authorcr {\tt\small \{meetshah, xinleic, mrf\}@fb.com, dparikh@gatech.edu} }

\date{}
\maketitle
\vspace*{-10mm}
\begin{abstract}

Despite significant progress in Visual Question Answering over the years, robustness of today's VQA models leave much to be desired. We introduce a new evaluation protocol and associated dataset (\vqarep) and show that state-of-the-art VQA models are notoriously brittle to linguistic variations in questions. 
\vqarep contains 3 human-provided rephrasings for 40k questions spanning 40k images from the VQA v2.0 validation dataset. 
As a step towards improving robustness of VQA models, we propose a model-agnostic framework that exploits cycle consistency. Specifically, we train a model to not only answer a question, but also generate a question conditioned on the answer, such that the answer predicted for the generated question is the same as the ground truth answer to the original question. 
Without the use of additional annotations,
we show that our approach is significantly more robust to linguistic variations than state-of-the-art VQA models, when evaluated on the \vqarep dataset. 
In addition, our approach outperforms state-of-the-art approaches on the standard VQA and Visual Question Generation tasks on the challenging VQA v2.0 dataset. 
\end{abstract}

\section{Introduction}
\label{intro_section}
\begin{figure}[!tbp]
\begin{center}
\includegraphics[width=0.46\textwidth]{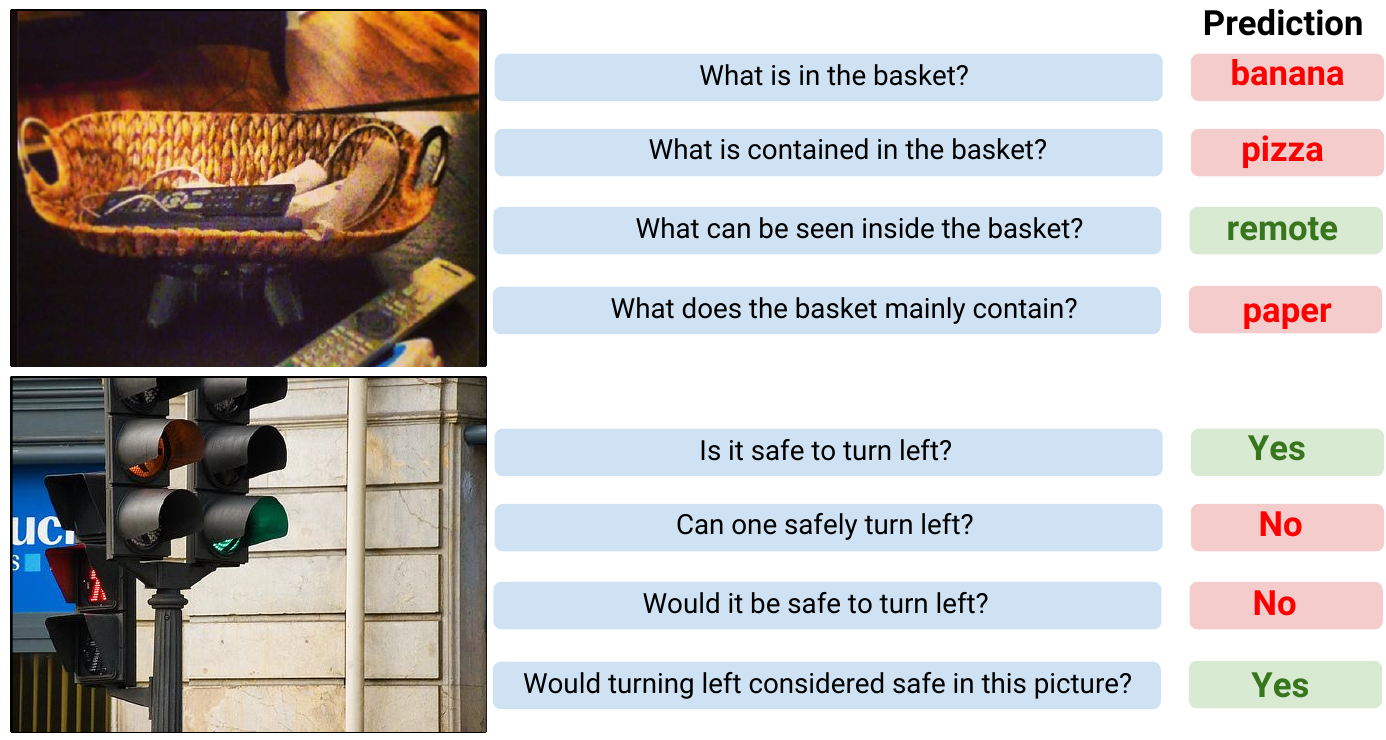}
\end{center}
   \caption{\textbf{Existing VQA models are brittle}. Shown above are examples from our new large-scale \textbf{\vqarep} dataset that enables systematic evaluation of robustness of VQA models to linguistic variations in the input question. Also shown are answers predicted by a state-of-the-art VQA model \cite{pythia18arxiv}.
   We see that the model predicts different answers for different reasonable rephrasings of the same question. We propose a novel model-agnostic framework that exploits cycle consistency in question answering and question generation to make VQA models more robust, 
   without using additional annotation.
   Moreover, it outperforms state-of-the-art models on the standard VQA and Visual Question Generation tasks on the VQA v2.0 dataset.
   }

\label{fig:failure_vqa}
\end{figure}

Visual Question Answering (VQA) applications allow a human user to ask a machine questions about images -- be it a user interacting with a visual chat-bot or a visually impaired user relying on an assistive device. As this technology steps out of the realm of curated datasets towards real-world settings, it is desirable that VQA models be robust to and consistent across reasonable variations in the input modalities. While there has been significant progress in VQA over the years~\cite{agrawal2016analyzing, johnson2017inferring, agrawal2018don, goyal2017making, Kim2018, pythia18arxiv, anderson2018bottom, andreas2016learning}, today's VQA models are, however, far from being robust.

VQA is a task that lies at the intersection of language and vision. Existing works have studied the robustness and sensitiveness of VQA models to meaningful semantic variations in images~\cite{goyal2017making}, changing answer distributions~\cite{agrawal2018don} and adversarial attacks~\cite{xu2018fooling} to images. However, to the best of our knowledge, no work has studied the robustness of VQA models to linguistic variations in the input question. This is important both from the perspective of VQA being a benchmark to test multi-modal AI capabilities (do our VQA models really ``understand'' the question when answering it?) and for applications (human users are likely to phrase the same query in a variety of different linguistic forms). However, today's state-of-the-art VQA models are brittle to such linguistic variations as can be seen in Fig.~\ref{fig:failure_vqa}. 

One approach to make VQA models more robust is to collect a dataset with diverse rephrasings of questions to train VQA models. This requires additional human annotation and thus is not always scalable in real-world settings. 
Alternatively, an automatic approach that does not require additional human intervention but results in a VQA model that is more robust to linguistic variations observed in the natural language open-ended questions is desirable. 

We propose a novel model-agnostic framework that relies on cycle consistency to learn robust VQA models without requiring additional annotation. Specifically, we train the model to not just answer a question, but also to generate diverse, semantically similar variations of questions conditioned on the answer. We enforce that the answer predicted for a generated question matches the ground truth answer to the original question. In other words, the model is being trained to predict the same (correct) answer for a question and its (generated) rephrasing.

Advantages of our proposed approach are two fold. 
First, enforcing consistent correctness across diverse rephrasings allows models to generalize to unseen 
semantically equivalent variations of questions at test time. The model achieves this by generating linguistically diverse rephrasings of questions on-the-fly and training with these variations. 
Second, a model trained generatively to generate a valid question given a candidate answer and image has a stronger multi-modal understanding of vision and language. Questions tend to have less learnable biases~\cite{liu2017ivqa}. As a result, models that can jointly perform the task of question generation and question answering are less prone to taking ``shortcuts'' and exploiting linguistic priors in questions. Indeed, we find that models trained with our approach outperform existing state-of-the-art models on both VQA and Visual Question Generation (VQG) tasks on VQA v2.0~\cite{goyal2017making}.

We also observed that one reason for limited development of VQA models robust to linguistic variations in input questions is due to the lack of a benchmark to measure robustness. 
A lack of such a benchmark makes it hard to quantitatively realize the inflated capabilities and limited multi-modal understanding of modern VQA models and consequently inhibits progress in pushing the state-of-the-art in multi-modal understanding aspects of computer vision. 
To enable quantitative evaluation of robustness and consistency of VQA models across linguistic variations in input questions, we collect a large-scale dataset -- \textbf{\vqarep} (Section \ref{sec:dataset}) based of the VQA v2.0 dataset~\cite{goyal2017making}. \vqarep contains 3 human-provided rephrasings for $\sim$40k questions on $\sim$40k images from the validation split of the VQA v2.0 dataset. We also propose metrics to measure the robustness of VQA models across different question rephrasings. Further, we benchmark several state-of-the-art VQA models~\cite{anderson2018bottom, ben2017mutan, Kim2018, pythia18arxiv} on our proposed \vqarep dataset to highlight the fragility of VQA models to question rephrasings. We observe a significant drop when VQA models are required to be consistent in addition to being correct (Section \ref{sec:results}), which reinforces our belief that existing VQA models do not understand language "enough". We show that VQA models trained with our approach are significantly more robust across question rephrasings than their existing counterparts on the proposed \vqarep dataset. \\

In this paper, our contributions are the following:
\begin{packed_itemize}
    \item We propose a model-agnostic cycle-consistent training scheme that enables VQA models to be more robust to linguistic variations observed in natural language open-ended questions. 
    \item To evaluate the robustness of VQA models to linguistic variations, we introduce a large-scale \textbf{\vqarep} dataset and an associated consensus score. \vqarep consists of 3 rephrasings for $\sim$40k questions on $\sim$40k images from the VQA v2.0 validation dataset, resulting in a total of $\sim$120k questions rephrasing by humans.
    \item  We show that models trained with our approach outperform state-of-the-art on the standard VQA and Visual Question Generation tasks on the VQA v2.0 dataset and are significantly more robust to linguistic variations on \vqarep.
\end{packed_itemize}

\begin{figure*}
\begin{center}
\includegraphics[width=\textwidth]{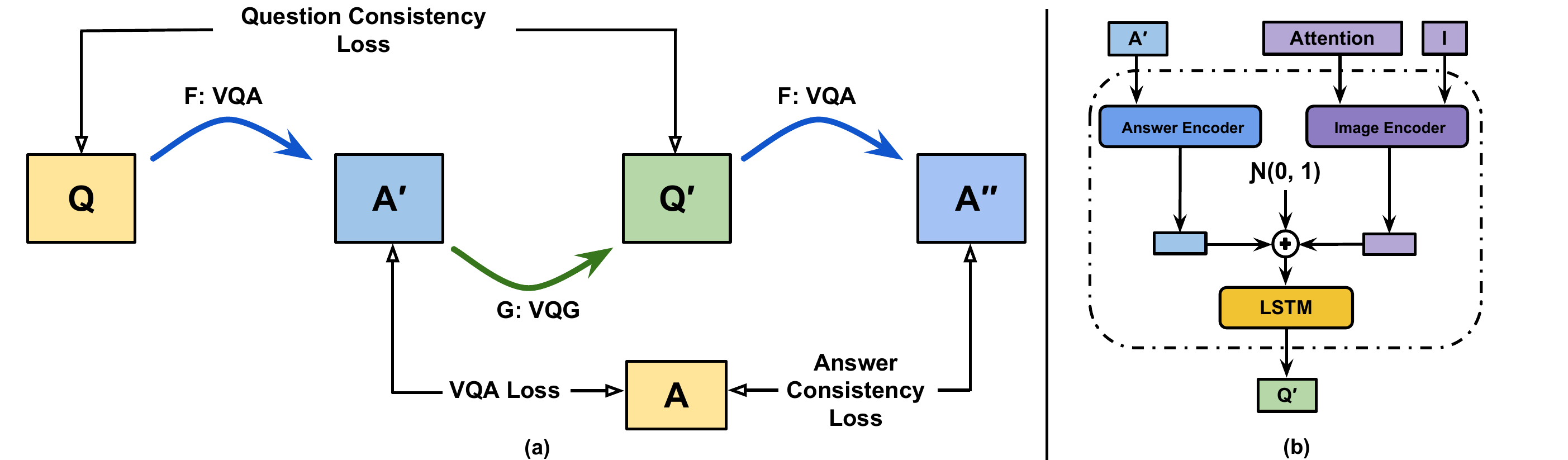}
\end{center}
   \caption{(a) \textbf{Abstract representation of the proposed cycle-consistent training scheme:} Given a triplet of image $I$, question $Q$, and ground truth answer $A$, a VQA model is a transformation $F:(Q,I)\mapsto A^\prime$ used to predict the answer $A^\prime$. Similarly, a VQG model $G:(A^\prime,I)\mapsto Q^\prime$ is used to generate a rephrasing $Q^\prime$ of $Q$. The generated rephrasing $Q^\prime$ is passed through $F$ to obtain $A^{\prime\prime}$ and consistency is enforced between $Q$ and $Q^\prime$ and between $A^{\prime}$ and $A^{\prime\prime}$. Image $I$ is not shown for clarity.  
   (b) \textbf{Detailed architecture of our visual question generation module $G$}. The predicted answer $A^\prime$ and image $I$ are embedded to a lower dimension using task-specific encoders and the resulting feature maps are summed up with additive noise and fed to an LSTM to generate questions rephrasings $Q^\prime$.}
\label{fig:model_arch}
\end{figure*}

\section{Related Work}
\textbf{Visual Question Answering.}
There has been tremendous progress in building models for VQA using LSTMs \cite{hochreiter1997long} and convolutional networks \cite{lecun1998gradient}. VQA models spanning different paradigms like attention networks \cite{yang2016stacked, Kim2018}, module networks \cite{hu2017learning, andreas2016learning, johnson2017inferring}, relational networks \cite{perez2017film} and multi-modal fusion \cite{ben2017mutan} have been proposed. Our method is model-agnostic and is applicable with any existing VQA architecture.

\textbf{Robustness.} 
Robustness of VQA models has been studied in several contexts \cite{agrawal2018don, xu2018fooling, goyal2017making}. For example, \cite{agrawal2018don} studies the robustness of VQA models to changes in the answer distributions across training and test settings; \cite{yinyang} analyzes the extent of visual grounding in VQA models by studying robustness of VQA models to meaningful semantic changes in images; \cite{xu2018fooling} shows that despite the use of an advanced attention mechanism, it is easy to fool a VQA model with very minor changes in the image. Our work, however, aims to complete the study in robustness by benchmarking and improving robustness of VQA models to linguistic and compositional variations in questions in the form of rephrasings. Robustness has also been studied in natural language processing (NLP) systems~\cite{ettinger2017towards, robusthobbs1992} in contexts of bias \cite{robuststede1992search, robustspranger2016}, domain-shift \cite{li2017robust} and syntactic variations \cite{robustiyyer2018adversarial}. To counter these issues in NLP systems, solutions like linguistically motivated data-augmentation \cite{li2017robust} and adversarial training \cite{robustiyyer2018adversarial} have been proposed. We study this in the context of visual question answering which is a multi-modal task which grounds language into the visual world.

\textbf{(Visual) Question Generation.} 
Question Generation (QG) as a task has been studied extensively by~\cite{ali2010automation, kalady2010natural, serban2016generating, Wangqgnet} in NLP. Generating questions conditioned on an image was introduced in \cite{mora2016towards} and a large-scale VQG dataset was collected by \cite{mostafazadeh2016generating} to evaluate visually grounded question generation capabilities of models. More recently, there has been work on generating questions that are diverse~\cite{jain2017creativity, yang2016stacked}. 
Training models to ask informative questions about an image in an active learning fixed-budget setting was explored in~\cite{misra2017learning}. 
While these techniques generate questions about an image in an answer-agnostic manner, techniques like \cite{liu2017ivqa} propose a variational LSTM based model trained with reinforcement learning to generate answer-specific questions for an image. More recently, \cite{li2018visual} generates answer-specific questions for specific question-types by modelling question generation as a dual task of question answering. Unlike \cite{li2018visual}, our method is not restricted to generating questions only for specific question types. Different from previous works, the goal of our VQG component is to automatically generate question rephrasings that make the VQA models more robust to linguistic variations. To the best of our knowledge, we are the first to demonstrate that the VQG module can be used to improve VQA accuracy in a cycle-consistent setting.

\vspace*{-3pt}

\textbf{Cycle-Consistent Learning.}
Using cycle-consistency to regularize the training of models has been used extensively in object tracking \cite{tracking}, machine translation \cite{translation}, unpaired image-to-image translation \cite{CycleGAN2017} and text-based question answering \cite{tang2018learning}. Consistency enables learning of robust models by regularizing transformations that map one interconnected modality or domain to the other. While cycle consistency has been used vastly in the domains involving a single modality (text-only or image-only), it hasn't been explored in the context of multi-modal tasks like VQA.
Cycle-consistency in VQA can be also thought of as an online data-augmentation technique where the model is trained on several generated rephrasings of the same question.

\section{Approach} 
\label{sec:method}
We now introduce our cycle-consistent scheme to train robust VQA models. Given a triplet of image $I$, question $Q$, and ground truth answer $A$, a generic VQA model can be formulated as a transformation $F:(Q,I)\mapsto A^\prime$, where $A^\prime$ is the answer predicted by the model as in Fig.~\ref{fig:model_arch}(a). Similarly, a generic VQG model can be formulated as a transformation $G:(A,I)\mapsto Q^\prime$ as in Fig.~\ref{fig:model_arch}(b). 
For a given $(I,Q,A)$ triplet, we first obtain an answer prediction $A^\prime$ using the VQA model $F$ for the original question $Q$. 
We then use the predicted answer $A^\prime$ and the image $I$ to generate a question $Q^\prime$ which is semantically similar to $Q$ using the VQG model $G$.
Lastly, we obtain a answer prediction $A^{\prime\prime}$ for the generated question $Q^\prime$.

Our design of consistency components is inspired by two beliefs.
Firstly, a model which can generate a semantically and syntactically correct question given a answer and an image, has a better understanding of the cross-modal connections among the image, the question and the answer, which make them a valid $(I, Q, A)$ triplet. 
Secondly, assuming the generated question $Q^\prime$ is a valid rephrasing of the original question, a robust VQA model should answer this rephrasing with the same answer as the original question $Q$.
In practice, however, there are several challenges that inhibit enforcement of cycle-consistency in VQA. We discuss these challenges and describe the key components of our framework geared to tackle them in the following sections.

\subsection{Question Generation Module}
\label{sec:vqg}
Since VQA is a setting where there is high disparity in the information content of involved modalities (a question and answer pair is a very lossy compressed representation of the image), learning transformations that map one modality to another is non-trivial. In cycle-consistent models dealing with single-modalities, transformations need to be learned across different domains of the same modality (image or text) with roughly similar information contents. However in a multi-modality transformation like VQG, learning a transformation from a low information modality (such as answer) to high information modality (question) needs additional supervision.
We provide this additional supervision to the VQG model in the form of attention. To generate a rephrasing $Q^\prime$, the VQG is guided to attend at regions of the image which were used by the VQA model to answer the original question $Q$. Unlike \cite{li2018visual}, this enables our models to generate questions \meetnew{more similar} to the original question from answers like ``yes", which could possibly have a large space of plausible questions.

We model the question generation module $G$ in a fashion similar to a conditional image captioning model. The question generation module consists of two linear encoders that transform attended image features obtained from VQA model and the distribution over answer space to lower dimensional feature vectors. We sum these feature vectors with additive noise and pass them through an LSTM which is trained to reconstruct the original question and optimized by minimizing the negative log likelihood with teacher-forcing. Note that unlike \cite{liu2017ivqa, li2018visual} we do not pass the one-hot vector representing the answer obtained, or an embedding of the answer obtained to the question generation, but rather the predicted distribution over answers. This enables the question generation module to learn to map the model's confidence over answers to the generated question. 

\meetnew{
Throughout the paper, \textbf{Q-consistency} implies addition of a VQG module $G$ on top of the base VQA model $F$ to generate rephrasings $Q^\prime$ from the image $I$ and the predicted answer $A^\prime$ with an associated Q-consistency loss $\mathcal{L}_{G}(Q, Q^\prime)$. 
Similarly, \textbf{A-consistency} implies passing all questions generated $Q^\prime$  by the VQG Model $G$ to the VQA model $F$ and an associated A-consistency loss $\mathcal{L}_{cycle}(A, A^{\prime\prime})$.
}
The overall loss can be written as:
\begin{equation}
\label{eq:total_loss}
\begin{split}
    \mathcal{L}_{total} = \mathcal{L}_{F}(A, A^\prime) + \lambda_{G}  \mathcal{L}_{G}(Q, Q^\prime) \\ +  \lambda_{C}  \mathcal{L}_{cycle}(A, A^{\prime\prime})
\end{split}
\end{equation} 
where $\mathcal{L}_{F}(A, A^\prime)$ and $\mathcal{L}_{cycle}(A, A^{\prime\prime})$ \meetnew{(\ie A-Consistency Loss)} are cross-entropy losses, $\mathcal{L}_{G}(Q, Q^\prime)$ \meetnew{(\ie Q-Consistency Loss)} is sequence generation loss \cite{decanlp} and $\lambda_{G}$, $\lambda_{C}$ are tunable hyperparameters. 

\begin{figure*}[htbp]
\begin{center}
\includegraphics[width=1.0\textwidth]{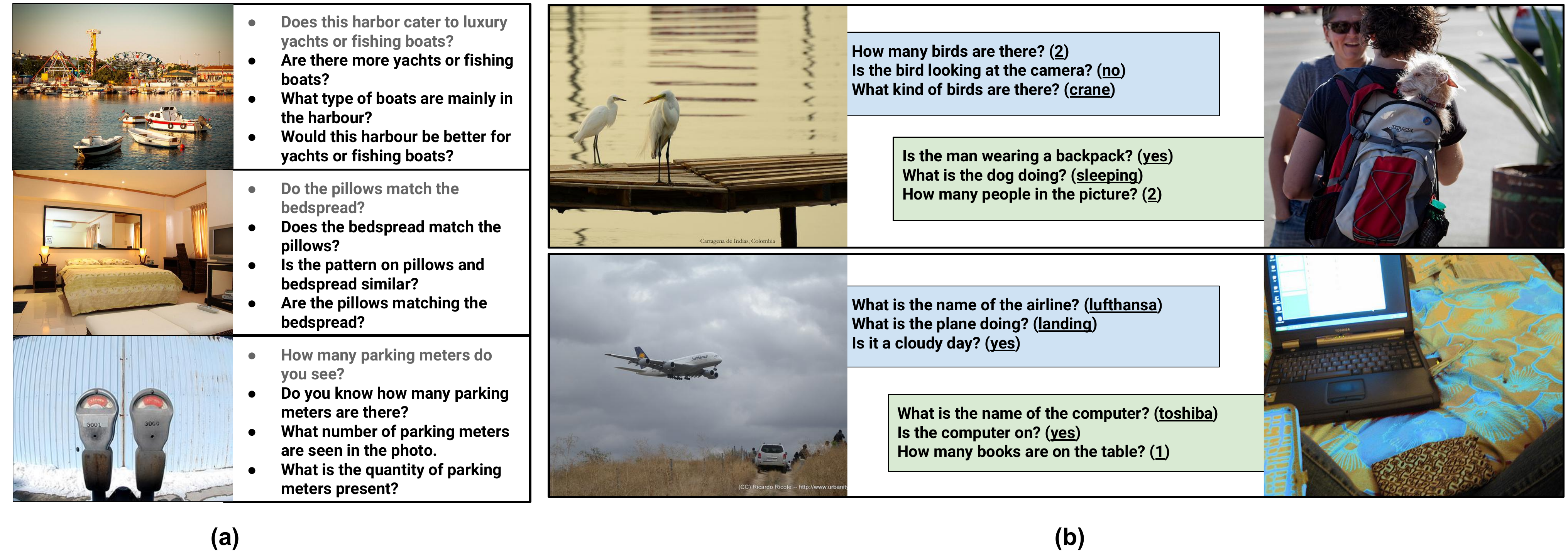}
\end{center}
   \caption{
   \textbf{ (a) Qualitative examples from our \vqarep dataset}.
   The first question (shown in gray) in each block is the original question from VQA v2.0 validation set, 
   the questions that follow (shown in black) are rephrasings collected in \vqarep.
   \textbf{(b) Qualitative examples of answer conditioned question generation (\underline{input answer}) by our VQG module
   }
   }
\label{fig:vqarep-dataset}
\end{figure*}
\subsection{Gating Mechanism}
One of the assumptions of our proposed cycle-consistent training scheme is that the generated question is always semantically and syntactically correct. However, in practice this is not always true. 
Previous attempts \cite{kafle2017data} at naively generating questions conditioned on the answer and using them without filtering to augment the training data have been unsuccessful.
Like the visual question answering module, the visual question generation module is also not perfect. Therefore not all questions generated by the question generator are coherent and consistent with the image, the answer and the original question.
To overcome this issue, we propose a gating mechanism, which automatically filters undesirable questions generated by the VQG model before passing them to the VQA model \meetnew{for A-consistency. The gating mechanism is only relevant when used in conjunction with A-consistency.}
We retain only those questions which either the VQA model $F$ can answer correctly or have a cosine similarity with the original question encoding greater than a threshold $T_{sim}$.
\subsection{Late Activation}
One key component of designing cycle consistent models is to prevent mode collapse. Learning cycle-consistent models in complex settings like VQA needs a carefully chosen training scheme. Since cycle-consistent models have several interconnected sub-networks learning different transformations, it is important to ensure that each of these sub-networks are working in harmony.
For example, if the VQA model $F$ and VQG model $G$ are jointly trained and consistency is enforced in early stages of training, it is possible that both models can just ``cheat" by both producing undesirable outputs. 
We overcome this by activating cycle-consistency at later stages of training, to make sure both VQA and VQG models have been sufficiently trained to produce reasonable outputs. Specifically, we enable the loss associated with cycle-consistency after a fixed $A_{iter}$ iterations in the training process.

We find these design choices for question generation module, gating mechanism and late activation to be crucial for effectively training our model.
We demonstrate this empirically via ablation studies in Table~\ref{vqa_table}.
As we want to increase the robustness of the VQA model to all generated variations, the weights between VQA models which answer the original question and the generated rephrasing are shared. %
Our formulation of cycle-consistency in VQA can be also thought of as an online data-augmentation technique where the model is trained on several generated rephrasings of the same question and hence is more robust to such anomalies during inference. We show that with clever training strategy, coupled with attention and carefully chosen model architectures for question generation, incorporating cycle consistency for VQA is possible and not only leads to models that are better performing, but also more robust and consistent. In addition, we show that this robustness also imparts VQA models the ability to better predict their own failures.

\section{\vqarep Dataset}
\label{sec:dataset}

In this section, we introduce the \vqarep dataset, which is the first dataset that enables evaluation of VQA models for robustness and consistency to different rephrasings of questions with the same meaning. 

We use the validation split of VQA v2.0 ~\cite{goyal2017making} as our base dataset which contains a total of 214,354 questions spanning over 40,504 images. We randomly sample 40,504 questions (one question per image) from the base dataset to form a sampled subset. We collect 3 rephrasings of each question in the sampled subset using human annotators in two stages. In the first stage, humans were primed with the original question and the corresponding true answer and asked to rephrase the question such that answer to the rephrased question remains the same as the original answer. To ensure rephrasings from first stage are \textit{syntactically} correct and \textit{semantically} inline with the original question, we filter the collected responses in the next stage. 

In the second stage, humans were primed with the original question and it's rephrasing and were asked to label the rephrasing invalid if:
(a) the plausible answer to the original question and it's rephrasing is different (\ie if the question and it's rephrasing have different intents) or 
(b) if the rephrasing is grammatically incorrect. 
We collected 121,512 rephrasings from the original 40504 questions in the first stage. Of these, 1320 rephrasings were flagged as invalid in the second stage and were rephrased again in the first stage. Humans were shown examples of incorrect rephrasings in the first stage to minimize the number of invalid rephrasings.

The final dataset consists of 162,016 questions (including the original 40,504 questions) spanning 40,504 images with an average of $\sim$3 rephrasings per original question. A few qualitative examples from the collected dataset can be seen in Fig.~\ref{fig:vqarep-dataset}(a). Additional details about the data collection, interfaces used and exhaustive dataset statistics can be found in Appendix~\ref{ss:dataset_sm}.
\textbf{Consensus Score.} Intuitively, for a VQA model to be consistent across various rephrasings of the same question, the answer to all rephrasings should be the same. We measure this by a Consensus Score $CS(k)$. For every group $Q$ consisting of $n$ rephrasings, we sample all subsets of size $k$.
The consensus score $CS(k)$ is defined as the ratio of the number of subsets where \emph{all} the answers are correct and the total number of subsets of size $k$. The answer to a question is considered correct if it has a non-zero VQA Accuracy $\theta$ as defined in~\cite{agrawal2016analyzing}.
CS(k) is formally defined as:
\begin{equation}
\label{eq:consesus_score}
CS(k) \quad = \sum_{Q^\prime \subset Q, |Q'|=k} \frac{\mathcal{S}(Q^\prime)}{\Mycomb{k}}
\end{equation}

\begin{equation}
\label{eq:all_correct}
\mathcal{S}(Q^\prime) = \begin{cases}
        \quad  1 & \text{if } \enskip \forall q \in Q^\prime \enskip \theta(q) > 0, \\
        \quad  0 & \text{otherwise}.
\end{cases}
\end{equation}

Where $\Mycomb{k}$ is number of subsets of size $k$ sampled from a set of size $n$.
As consensus score is a all-or-nothing score, to achieve a non-zero consensus score at $k$ for a group of questions $Q$, the model has to answer at least $k$ questions correctly in a group of questions $Q$. When $k=|Q|$ (\eg when $k=4$ in \vqarep), the model needs to answer all rephrasings of a question and the original question correctly in order to get a non-zero consensus score. It is evident that a model with higher average consensus score at high values of $k$ is quantitatively more robust to linguistic variations in questions than a model with a lower score.

\section{Experiments}
\label{sec:results}

\begin{table}[t]\centering
\setlength{\tabcolsep}{3.5pt}
\begin{tabular}{lcccc|cc}
 \toprule
  \textbf{Model} & \multicolumn{4}{c}{\textbf{CS(k)}} & \multicolumn{2}{c}{\textbf{VQA Accuracy}} \\
  ~ & \textbf{k=1} & \textbf{k=2} & \textbf{k=3} & \textbf{k=4} & \textbf{ORI} & \textbf{REP} \\
  
 \midrule
 MUTAN~\cite{ben2017mutan}       & 56.68& 43.63	& 38.94	& 32.76	& 59.08	& 46.87 \\ 
 \midrule
 BUTD ~\cite{anderson2018bottom} & 60.55	& 46.96	& 40.54	& 34.47	& 61.51	& 51.22 \\
 BUTD + CC                       & \textbf{61.66} &	\textbf{50.79} &	\textbf{44.68} &	\textbf{42.55} &	\textbf{62.44} &	\textbf{52.58} \\ 
 \midrule
 Pythia~\cite{pythia18arxiv}    & 63.43	& 52.03	& 45.94	& 39.49	& 64.08	& 54.20 \\
 Pythia + CC                   & \textbf{64.36}	& \textbf{55.45}	& \textbf{50.92}	& \textbf{44.30}	& \textbf{64.52}	& \textbf{55.65}\\
 \midrule
 BAN~\cite{Kim2018}              & 64.88	& 53.08	& 47.45	& 39.87	& 64.97	& 55.87 \\
 BAN + CC                       & \textbf{65.77}	& \textbf{56.94}	& \textbf{51.76}	& \textbf{48.18}	& \textbf{65.87}	& \textbf{56.59} \\ 
\bottomrule
\end{tabular}
  \vspace{2mm}
  \caption{\textbf{Consensus performance on \vqarep dataset}. CS(k) as defined in Eq.~\ref{eq:consesus_score} is consensus score which is non-zero only if \emph{at least} $k$ rephrasings are answered correctly, zero otherwise; averaged across all group of questions.
  ORI represent a split of questions from \vqarep which are original questions from VQA v2.0 and their corresponding rephrasings are represented by the split REP.
  Models trained with our cycle-consistent (CC) framework consistently outperform their baseline counterparts at all values of $k$.
  }
  \label{robustness_table}
\end{table}

\subsection{Consistency Performance}

We start by benchmarking a variety of existing VQA models on our proposed \vqarep dataset.

\textbf{MUTAN}~\cite{ben2017mutan}~\footnote{https://github.com/Cadene/vqa.pytorch
} parametrizes bilinear interactions between visual and textual representations using a multi-modal low-rank decomposition. MUTAN uses skip-thought \cite{kiros2015skip} sentence embeddings to encode the question and Resnet-152~\cite{he2016deep} to encode images. MUTAN achieves 63.20\% accuracy on VQA v2.0 test-dev. Among all models we analyze, MUTAN is the only model which uses sentence embeddings to encode questions and Resnet to encode images.

\textbf{Bottom-Up Top-Down Attention (BUTD)}~\cite{anderson2018bottom}~\footnote{https://github.com/hengyuan-hu/bottom-up-attention-vqa} incorporates bottom-up attention in VQA by extracting features associated with image regions proposed by Faster-RCNN~\cite{ren2015faster} pretrained on Visual Genome \cite{krishnavisualgenome}. BUTD model won the VQA Challenge in 2017 and achieves 66.25\% accuracy on VQA v2.0 test-dev.

\textbf{Pythia}~\cite{pythia18arxiv}~\footnote{https://github.com/facebookresearch/pythia} extends the BUTD model by incorporating co-attention \cite{lu2016hierarchical} between question and image regions. Pythia uses features extracted from Detectron \cite{Detectron2018} pretrained on Visual Genome. An ensemble of Pythia models won the VQA Challenge in 2018 using additional training data from Visual Genome \cite{krishnavisualgenome} and using additional Resnet\cite{he2016deep} features. In this study, we use Pythia models which do not use Resnet features. Pythia without using Resnet features, achieves an accuracy of 68.43 \% on VQA v2.0 test-dev.

\begin{table}[t]
\centering

\begin{tabular}{l|c|c} 
 \toprule
\textbf{Model} & \textbf{val} & \textbf{test-dev} \\
\midrule

MUTAN \cite{ben2017mutan}                         & 61.04 & 63.20 \\
\midrule
BUTD \cite{anderson2018bottom}                    & 65.05 & 66.25 \\ 
 + Q-consistency                                  & 65.38 & 66.83 \\
 \quad + A-consistency                            & 60.84 & 62.18 \\
 \quad\quad  + Gating                             & \textbf{65.53} & \textbf{67.55} \\
\midrule
Pythia \cite{pythia18arxiv}                  & 65.78 & 68.43 \\
+ Q-consistency                              & 65.39 & 68.58 \\
\quad + A-consistency                        & 62.08 & 63.77 \\
\quad\quad + Gating                          & \textbf{66.03} & \textbf{68.88} \\
\midrule

BAN \cite{Kim2018}                           & 66.04 & 69.64 \\ 
+ Q-consistency                              & 66.27 & 69.69 \\
\quad + A-consistency                        & 64.96 & 66.31 \\
\quad\quad + Gating                          & \textbf{66.77} & \textbf{69.87} \\
 \bottomrule
\end{tabular}
  \vspace{2mm}
  \caption{\textbf{VQA Performance and ablation studies on VQA v2.0 validation and test-dev splits}. Each row in blocks represents a component of our cycle-consistent framework added to the previous row. First row in each block represents the baseline VQA model $F$.
  Q-consistency implies addition of a VQG module $G$ to generate rephrasings $Q^\prime$ from the image $I$ and the predicted answer $A^\prime$ with an associated VQG loss $\mathcal{L}_{vqg}(Q, Q^\prime)$. 
  A-consistency implies passing all the generated questions $Q^\prime$ to the VQA model $F$ and an associated loss $\mathcal{L}_{cycle}(A, A^\prime)$.
  Gating implies the use of gating mechanism to filter undesirable generated questions in $Q^\prime$ and passing the remaining to VQA model $F$.
  Models trained with our cycle-consistent \meetnew{(last row in each block)} framework consistently outperform baselines.}
  \label{vqa_table} 
\end{table}

\textbf{Bilinear Attention Networks (BAN)~\cite{Kim2018}~\footnote{https://github.com/jnhwkim/ban-vqa}} combines the idea of bilinear models and co-attention \cite{lu2016hierarchical} between image regions and words in questions in a residual setting. Similar to \cite{anderson2018bottom}, it uses Faster-RCNN \cite{ren2015faster} pretrained on Visual Genome \cite{krishnavisualgenome} to extract image features. In all our experiments, for a fair comparison, we use BAN models which do not use additional training data from Visual Genome. BAN achieves the current state-of-the-art single-model accuracy of 69.64 \% on VQA v2.0 test-dev without using additional training data from Visual Genome.

\begin{figure*}
\begin{center}
\includegraphics[width=\textwidth]{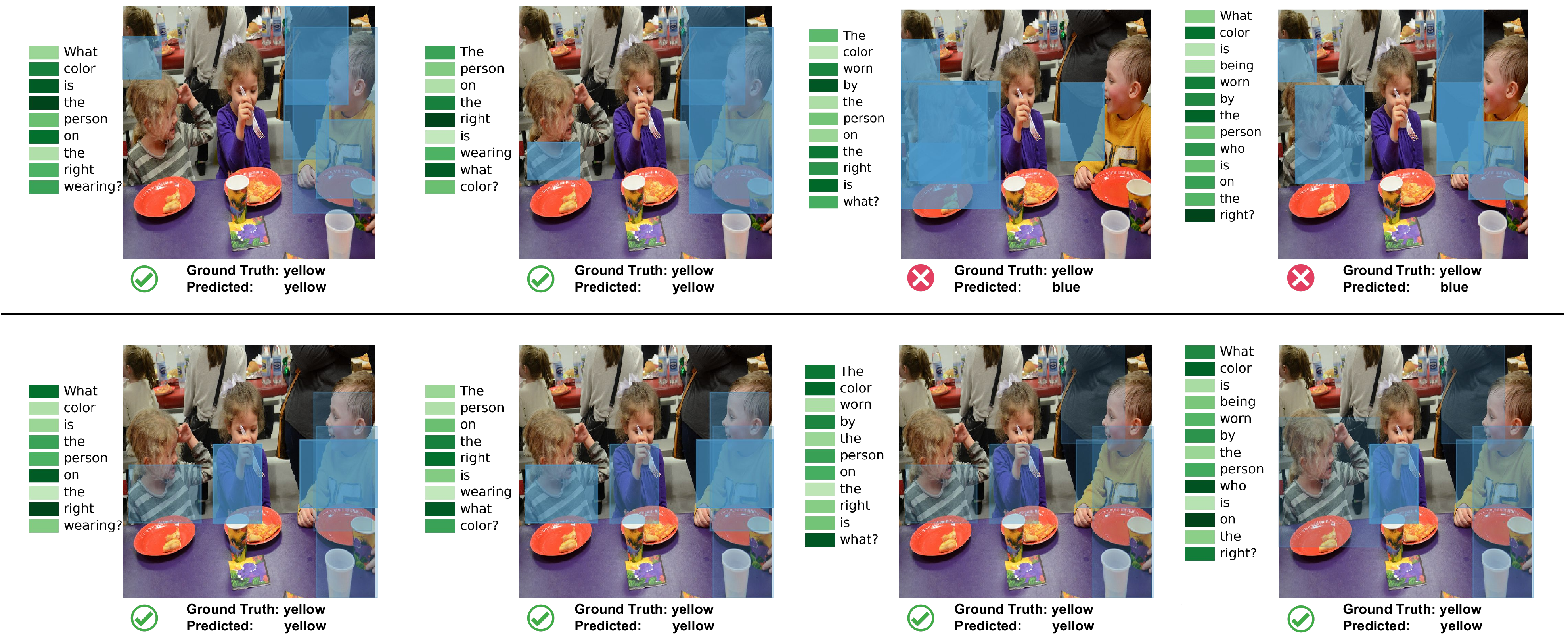}
\end{center}
   \caption{
   \textbf{Visualization of textual and image region attention across question variants:} The top row shows attention and predictions from a Pythia~\cite{pythia18arxiv} model, the bottom row shows attention and predictions from the same Pythia model, but trained using our cycle-consistent approach. Our model attends to relevant image regions for all rephrasings and answers them correctly. The baseline Pythia counterpart, however, fails to attend over relevant image regions for some rephrasings.
   }
\label{fig:attention_vis}
\end{figure*}

\begin{table*}[t]
\centering
\begin{tabular}{c|c|c|c|c|c|c|c} 
 \toprule
\textbf{ Model}                                         & \textbf{BLEU-1} & \textbf{BLEU-2} & \textbf{BLEU-3} & \textbf{BLEU-4} & \textbf{ROUGE-L} & \textbf{METEOR} & \textbf{CIDER} \\
\midrule
iQAN*~\cite{li2018visual}                      & 0.582 & 0.467 & 0.385 & 0.320 & 0.617 & 0.276 & 2.222 \\ 
Pythia + CC*                                          & \textbf{0.708} & \textbf{0.561} & \textbf{0.438} & \textbf{0.339} & \textbf{0.627} & \textbf{0.284} & \textbf{2.301} \\ \midrule
iVQA~\cite{liu2017ivqa}                        & 0.430 & 0.326 & 0.256 & 0.208 & 0.468 & 0.205 & 1.714 \\ 
Pythia + CC                                           & \textbf{0.486} & \textbf{0.368} & \textbf{0.287} & \textbf{0.226} & \textbf{0.556} & \textbf{0.225} & \textbf{1.843} \\
\bottomrule 
\end{tabular}
  \vspace{2mm}
  \caption{\textbf{Question Generation Performance on VQA v2.0 validation set}, * signifies results on a constrained subset as done in \cite{li2018visual}. CC represents models trained with our approach.}
\label{qg_table}
\end{table*}

\textbf{Implementation Details}
For all models trained with our cycle-consistent framework, we use the values $T_{sim}{=}0.9$, $\lambda_{G}{=}1.0$, $\lambda_{C}{=}0.5$ and $A_{iter}{=}5500$.
When reporting results on the validation split and \vqarep we train on the training split and when reporting results on the test split we train on both training and validation splits of VQA v2.0.
Note that we \emph{never} explicitly train on the collected \vqarep dataset and use it purely for evaluation purposes. We use publicly available implementations of each backbone VQA model.
The hidden size of the LSTM used in VQG module is 1024 and the linear encoders used to encode the answer and image in VQG have dimensions of 300 each.
Additional details about model-specific hyperparameters can be found in Appendix~\ref{ss:hyperparams}.

We measure the robustness of each of these models on our proposed \vqarep dataset using the consensus score (Eq.~\ref{eq:consesus_score}). Table~\ref{robustness_table} shows the consensus scores at different values of $k$ for several VQA models. We see that all models suffer significantly when measured for consistency across rephrasings. 
For \eg, the performance of Pythia (winner of 2018 VQA challenge) is reduced to a consensus score of 39.49\% at $k=4$.
Similar trends are observed for MUTAN, BAN and BUTD. The drop increases with increasing $k$, the number of rephrasings used to measure consistency. Models like BUTD, BAN and Pythia which use word-level encodings of the question suffer significant drops. It is interesting to note that even MUTAN which uses skip-thought based sentence encoding \cite{kiros2015skip} suffers a drop when checked for consistency across rephrasings. We observe that BAN + CC model trained with our proposed cycle-consistent training framework consistently outperforms its counterpart BAN and all other models at all values of $k$.

Fig~\ref{fig:attention_vis} qualitatively compares the textual and visual attention (over image regions) over 4 rephrasings of a question. The top row shows attention and predictions from a Pythia model, while the bottom row shows attention and predictions from the same Pythia model, but trained using our framework. Our model attends at relevant image regions for all rephrasings and answers all of them correctly. The Pythia counterpart, however, fails to attend over relevant image regions for some rephrasings and answers those rephrasings incorrectly. This qualitatively demonstrates the robustness of models trained with our framework. 

\subsection{Visual Question Answering Performance}
We now evaluate our approach and various ablations on the standard task of question answering on VQA v2.0 dataset~\cite{goyal2017making}. 
We compare the performance of several VQA models on the validation and test-dev splits of VQA v2.0. It consists of 443,757 training, 214,354 validation and 447,793 testing questions
spanning over 82,783, 40,504 and 81,434 images respectively. Table~\ref{vqa_table} shows the VQA scores of different models on validation and test-dev splits. We show that BUTD, Pythia and BAN models trained with our cycle-consistent framework outperform their corresponding baselines.

We show the impact of each component of our cycle-consistent framework by performing ablation studies on our models. We study the marginal effect of components like question consistency (Q-consistency), answer consistency (A-consistency) and gating mechanism by adding them step-by-step to the base VQA model $F$. 
Q-consistency implies addition of a VQG module $G$ to generate rephrasings $Q^\prime$ from the image $I$ and the predicted answer $A^\prime$ with an associated VQG loss $\mathcal{L}_{vqg}(Q, Q^\prime)$. As shown in Table~\ref{vqa_table}, we see that addition of question consistency slightly improves performance of each VQA model. Inline with observations in~\cite{li2018visual}, this shows that indeed models which can generate questions from the answer have better multi-modal understanding and in turn are better at visual question answering.
A-consistency implies passing all the generated questions $Q^\prime$ to the VQA model $F$ and an associated loss $\mathcal{L}_{cycle}(A, A^\prime)$. 
As seen in Table~\ref{vqa_table}, we see that naively passing all the generated questions to the VQA model $F$ leads to significant reduction in performance than the base model $F$. This goes in line with our earlier discussion that not all questions generated are \emph{valid} rephrasings of the original question and hence enforcing consistency between the answers of two invalid pairs of questions naturally leads to degradation in performance. 
Finally we show the effect of using our gating mechanism to filter undesirable generated questions in $Q^\prime$ and passing the remaining to VQA model $F$. We see that all VQA models perform \meetnew{consistently} better when using a gating than just using Q-consistency.

\meetnewn{We also experimented with Pythia model configurations where the VQG model uses unattended image features (unlike the default setting which uses image features with attention from the VQA model). 
We found that with this configuration, our approach still shows improved performance over the baseline. However, the question generation quality is relatively poor, and the overall gain is smaller (3.58\% in consistency $CS(k=4)$ and 0.2\% in VQA accuracy) compared to when using attention (8.08\% and 0.5\% respectively) -- likely because attention helps in generating more-focused rephrasings}

\subsection{Visual Question Generation Performance}
Recall that our model also includes a VQG component which generates questions conditioned on an answer and image. Since the overall performance of our framework relies highly on the performance of question generation module, we evaluate our VQG component performance as well on commonly used image captioning metrics. We compare our VQG component to several answer-conditional VQG models on the VQA v2.0 dataset. We use standard image captioning metrics CIDEr~\cite{cider}, BLEU~\cite{bleu}, METEOR~\cite{meteor} and ROUGE-L~\cite{rouge} as used in~\cite{liu2017ivqa}. We compare our approach to two recently proposed visual question generation approaches. \textbf{iVQA}~\cite{liu2017ivqa} uses a variational LSTM model trained with reinforcement learning to generate answer-specific questions for an image. Syntactic correctness, diversity and intent of the generated question are used to allocate rewards. 
\textbf{iQAN}~\cite{li2018visual} generates answer-specific questions by modelling question generation as a dual task of question answering and sharing parameters between question answering and question generation modules. Since iQAN can only generate a specific type of questions, for a fair comparison, we compare to iQAN only on a subset of the dataset containing questions from these specific types. As shown in Table~\ref{qg_table}, we observe that our question generation module trained with cycle-consistency consistently outperforms iVQA~\cite{liu2017ivqa} and iQAN~\cite{li2018visual} on all metrics. \meetnew{A few qualitative examples of answer conditioned questions generated by our VQG model can be seen in Fig.~\ref{fig:vqarep-dataset}(b).} Additional examples can also be found in the Appendix~\ref{ss:question_generation}.

\begin{table}[t]\centering

\begin{tabular}{l|c|c|c}
\toprule
 \textbf{Model} & \textbf{Precision} & \textbf{Recall} & \textbf{F1} \\
\midrule
 BUTD \cite{anderson2018bottom}    & 0.71 & 0.78 & 0.74 \\ 
 \quad + FP                        & \textbf{0.74} & \textbf{0.85} & \textbf{0.79} \\
 \midrule
 BUTD  + CC                                & 0.73 & 0.79 & 0.76 \\
 \quad\quad + FP                           & \textbf{0.78} & \textbf{0.83} & \textbf{0.80} \\
 \midrule
 Pythia \cite{pythia18arxiv}       & 0.74 & 0.79 & 0.76 \\ 
 \quad + FP                        & \textbf{0.76} & \textbf{0.88} & \textbf{0.82} \\
 \midrule
 Pythia + CC                       & 0.77 & 0.81 & 0.77 \\
 \quad\quad + FP                   & \textbf{0.82} & \textbf{0.84} & \textbf{0.83} \\
 \bottomrule

\end{tabular}
  \vspace{2mm}
  \caption{
  \textbf{Failure prediction performance on VQA v2.0 validation dataset}. 
  Each row in blocks represents a component added to the previous row. CC represents models trained with our cycle-consistent framework and FP represents models with an additional binary classification Failure Prediction submodule to predict if the predicted answer $A^\prime$ is correct given a question and image pair ($Q$, $I$). For models trained without the FP module, scores are obtained by thresholding the answer confidences.
  }
  \label{fp_table}
\end{table}

\vspace*{-1mm}
\subsection{Failure Prediction Performance} 
In previous results, we show that by training models to generate and answer questions while being consistent across both tasks leads to improvement in performance and robustness. Another way of testing robustness of these models is to see if models can predict their own failures. 
A robust model is less confident about an incorrect answer and vice versa. 
Motivated by this, we seek to verify if models trained with our cycle-consistent framework can identify their own failures \ie correctly identify if they're wrong about a prediction. 
\meetnew{To this end, we use two failure predictions schemes. First, we naively threshold the confidence of the predicted answer. All answers above a particular threshold are marked as correctly answered and vice versa. Second, we design a failure prediction binary classification module (FP), which predicts for a given image $I$, question $Q$ and answer $A^\prime$ (predicted by the base VQA model $F$), whether the predicted answer is correct for the given  $(I, Q)$ pair. The FP module uses image and answer encoders similar to those used in the question generation module (Section~\ref{sec:vqg}) and makes use of the question representation from the base VQA model as the question encoding. These encodings are concatenated and passed to a linear layer for binary classification. The FP module is trained keeping the parameters of the base VQA model frozen.}
In Table~\ref{fp_table}, we show the failure prediction performance of the baseline VQA models and models trained with our proposed framework. 
\meetnew{It shows that the cycle consistency framework, even \emph{without} an explicit failure predictor module, makes the models more calibrated -- more capable of detecting their own failures.
In both settings: (a) when using naive confidence thresholding (not marked as ``+ FP'' in the Table) and (b) using a specifically designed submodule to detect failures (marked as ``+ FP''), models trained with our cycle-consistent training framework are better than their corresponding baselines. 
We see similar improvments in detecting failures for both BUTD and Pythia models, which shows that our cycle-consistency framework is model agnostic.}
This also shows that not only does cycle-consistent training make models robust to linguistic variations, but also allows them to be aware of their failures.

\vspace*{-2mm}
\section{Conclusion}
In this paper, we propose a novel model-agnostic training strategy to incorporate cycle consistency in VQA models to make them robust to linguistic variations and self-aware of their failures. 
We also collect a large-scale dataset, \vqarep and propose a consensus metric to measure robustness of VQA models to linguistic variations of a question. 
We show that models trained with our training strategy are robust to linguistic variations, and achieve state-of-the-art performance in VQA and VQG on VQA v2.0 dataset.

\newpage

{\small
\bibliographystyle{ieee}
\bibliography{egbib}
}
\clearpage
\appendix

\makeatletter
\def\@maketitle
   {
   \newpage
   \null
   \vskip .05in
   \begin{center}
      {\Large \bf \@title \par}
   \end{center}
   }
\makeatother

\date{}

The appendix is organized as follows:
\begin{packed_itemize}
    \item Section~\ref{ss:dataset_sm} covers information about the dataset collection pipeline, user interface and provides some dataset statistics.
    \item Section~\ref{ss:attention} shows qualitative examples of how attention over image regions varies for VQA models when different rephrasings of the same question are used as input. 
    
    \item Section~\ref{ss:attention_con} describes an attention based consistency strategy that we experimented with, but did not improve performance (and so was not a part of our final model presented in the paper).
    \item Section~\ref{ss:question_generation} shows qualitative examples of answer conditioned questions generated by our VQG module.
    \item Section~\ref{ss:hyperparams} lists the hyperparameters used for each base VQA model.
\end{packed_itemize}

\section{Dataset Details}
\label{ss:dataset_sm}

\begin{figure}[htbp]
\begin{center}
\includegraphics[width=1.0\linewidth]{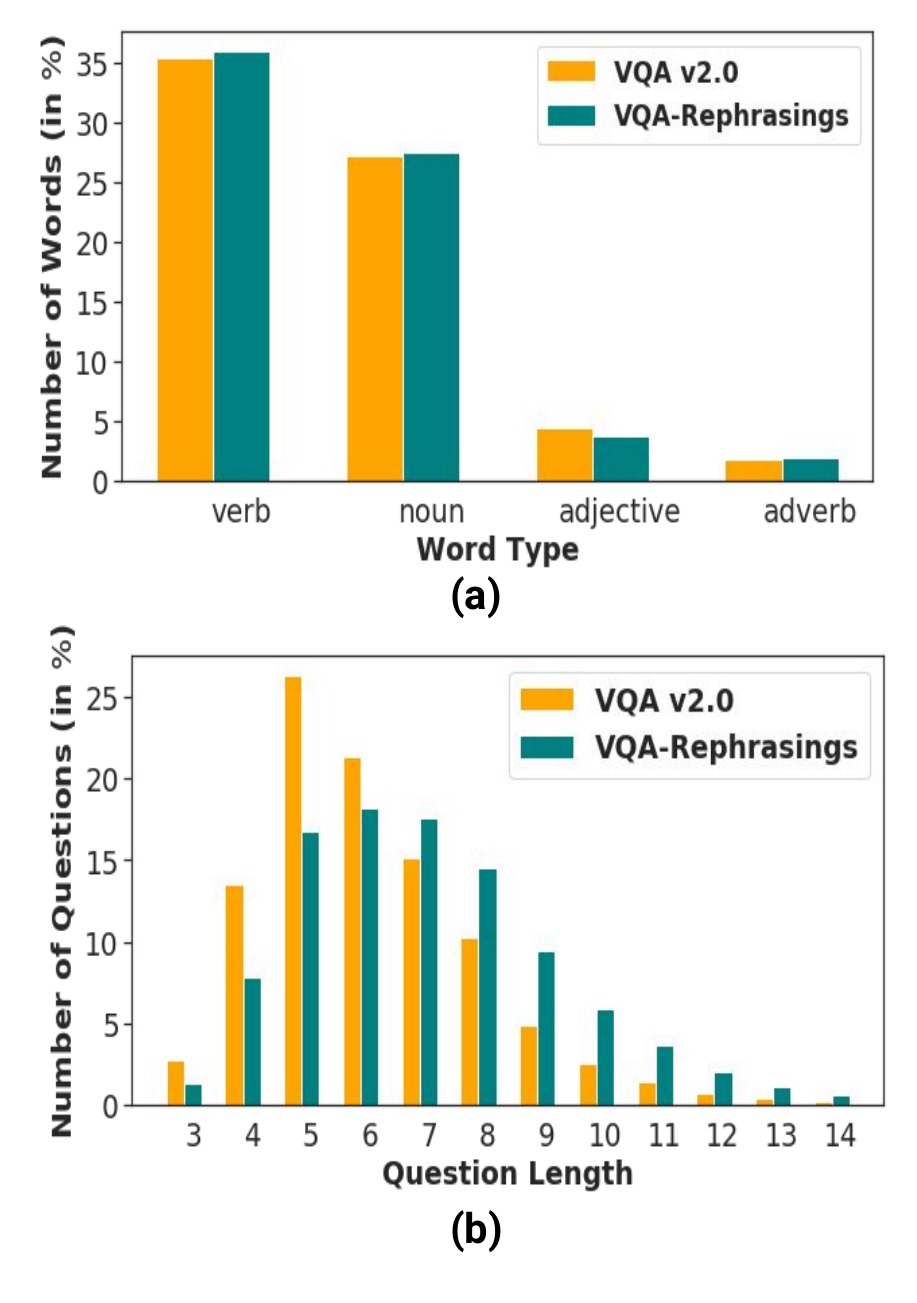}
\end{center}
   \caption{{\bf Dataset Statistics.}
   (a) Shows the number of words (in percentage) belonging to different Parts-of-Speech tags. The distributions follow similar trends in \vqarep and VQA v2.0.  
   (b) Shows the number of questions (in percentage) with varying lengths. The average length of questions in \vqarep is 7.15 which is slightly higher than the average length in VQA v2.0, which is 6.32.
   }
   
\label{fig:vqarep_stats}
\end{figure}

\textbf{Statistics.}
    Fig~\ref{fig:vqarep_stats}(a) shows the number of words (in percentage) belonging to different Parts-of-Speech tags. The distributions follow almost similar trends in \vqarep and VQA v2.0. This shows that the rephrasings are not obtained by merely adding more adjectives or adverbs in the original question.
    Fig~\ref{fig:vqarep_stats}(b) shows the number of questions (in percentage) with varying lengths. The average length of questions in \vqarep is 7.15 which is slightly higher than the average length in VQA v2.0, which is 6.32.

\textbf{Interface.}
    \meetnew{We used a simplistic web interface to} collect rephrasings from human annotators. The interface provided three examples of invalid rephrasings and their corresponding explanations to help human annotators understand the task better. We A/B tested with 50 questions using all 4 combinations of:
    \begin{packed_itemize}
        \item Showing both valid and invalid rephrasing examples and explanations.
        \item Showing only valid and no invalid rephrasing examples and explanations.
        \item Showing none of valid and invalid rephrasing examples and explanations.
        \item Showing no valid and only invalid rephrasing examples and explanations.
    \end{packed_itemize}
    We found (via manual inspection) that the last setup provided higher quality data, and used that as our final interface \meetnew{choice}.

\begin{figure*}[htbp]
\begin{center}
\includegraphics[width=1.0\textwidth]{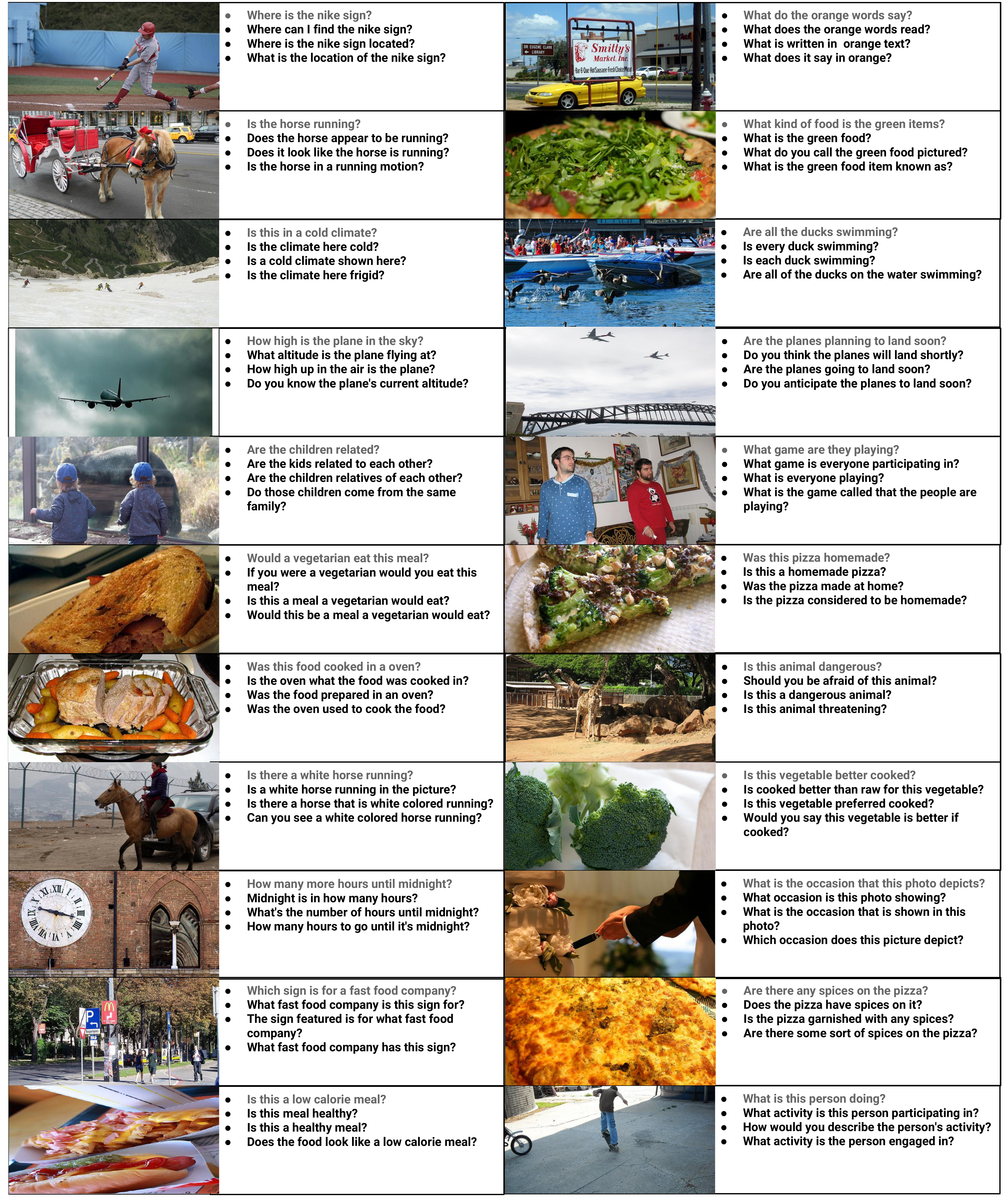}
\end{center}
   \caption{
   \textbf{Examples from our \vqarep dataset}.
   The first question (shown in gray) in each block is the original question from VQA v2.0 validation set, 
   the questions that follow (shown in black) are rephrasings collected in \vqarep.
   }
\label{fig:vqarep-dataset-sm}
\end{figure*}

\textbf{Examples.}
    Fig~\ref{fig:vqarep-dataset-sm} shows several qualitative examples from the \vqarep dataset. We see that the rephrasings  maintain the intent of the original question while varying linguistically.

\section{Attention Analysis}
\label{ss:attention}
\begin{figure*}
\begin{center}
\includegraphics[width=\textwidth]{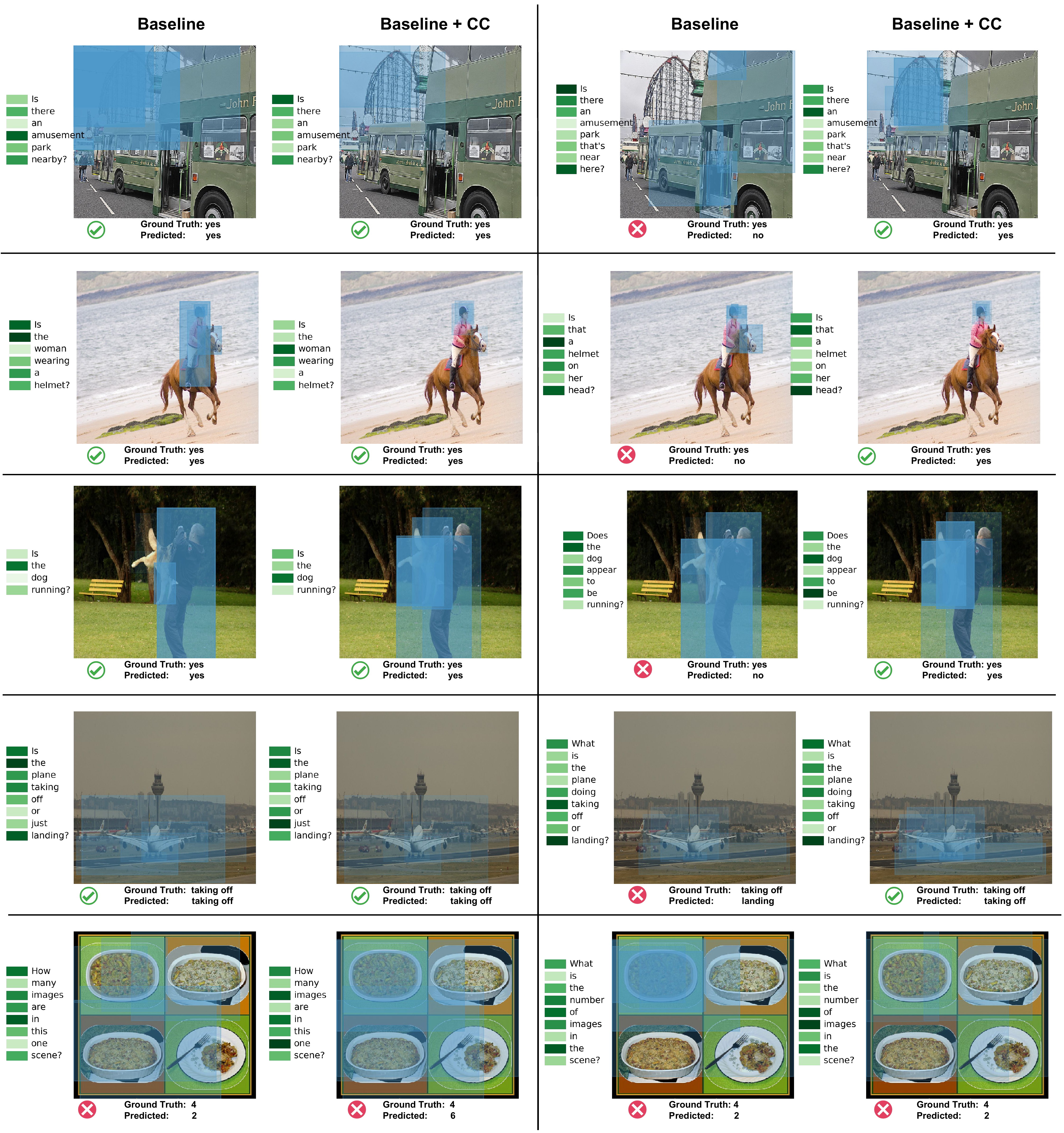}
\end{center}
   \caption{
    Visualization of textual and image region attention for different question variants: 
    Each row compares answers predicted and attention for two question rephrasings using a baseline Pythia~\cite{pythia18arxiv} model and the same Pythia model trained with our framework (Pythia + CC). 
    Higher opaqueness in highlighted regions represents higher attention.
    First and third rows show the output of a Pythia model (baseline) and second and forth rows show the output of a Pythia model (baseline + CC) trained with our framework.  
    As one can see, in most examples, the attention over image regions doesn't vary much for models trained with our framework.
    However for the baseline model, one can see that by very minor linguistic changes in the question it is possible to predict completely different answers (Row 2, Columns 1 and 3). 
    These examples qualitatively demonstrate the robustness of models trained with our framework.
   }
\label{fig:attention_vis_sm}
\end{figure*}

    Fig~\ref{fig:attention_vis_sm} qualitatively compares the textual and visual attention (over image regions) for rephrasings of a question.
    Each row compares predicted answers and attention from a baseline Pythia~\cite{pythia18arxiv} model and the same Pythia model trained with our framework (Pythia + CC), using two question rephrasings.
    First and third row shows the outputs of a Pythia model (baseline) and second and forth row shows the output of a Pythia model (baseline + CC) trained with our framework.  
    We see that in most examples, the attention over image regions doesn't vary across rephrasings for models trained with our framework (and the model answers the questions correctly). However for the baseline model, one can see that minor linguistic changes in the question can result in completely different answers (Row 2, Columns 1 and 3). This qualitatively demonstrates the robustness of models trained with our framework.
    Since the baseline Pythia model doesn't include a counting module, it doesn't perform well on questions requiring counting. As a result we see that both the baseline and its cycle-consistent counterpart perform poorly on counting questions (Row 5, Columns 1 through 4).

\section{Attention Consistency}
\label{ss:attention_con}
    Intuitively, it seems like training the VQA model to attend over the same image regions for different rephrasings of a question should improve the robustness of the model. We tried to enforce this in our cycle-consistent framework using an additional attention consistency loss.
    
    Recall that for a given image $I$, question $Q$ and answer $A$, our model consists of a VQA model $F$ which takes ($Q$, $I$) as an input and uses the question to attend over image regions with attention $\gamma_{Q}$ and predicts an answer $A^\prime$. We also have a VQG model $G$ which uses the predicted answer $A^\prime$ and image $I$ to generate a question $Q^\prime$. Intuitively, the VQA model should attend over the same image regions when answering $Q^\prime$. In other words, the attention over image regions $\gamma_{Q^\prime}$ used by the VQA model to answer $Q^\prime$  should be close to the $\gamma_{Q}$. We added an additional attention consistency loss to the total loss which reduces the $L_{2}$ norm between these two attentions.
    
    However, we found that this leads to reduction in model performance. Specifically, this reduces the performance of a cycle consistent Pythia model by 1.34\% VQA accuracy when evaluated on the VQA v2.0 validation split (training on train split only). 

    We suspect one reason why enforcing attention consistency across rephrasings reduces performance is perhaps because minimizing a large number of diverse losses ( cross entropy losses $\mathcal{L}_{F}$ and $\mathcal{L}_{cycle}$ for VQA, sequence generation loss $\mathcal{L}_{G}$ for VQG and mean squared loss $\mathcal{L}_{attention}$ for attention consistency) is a hard problem to optimize.
    Concretely identifying why enforcing attention consistency across question rephrasings hurts performance is currently under investigation and is part of future work. We find naively matching attentions across question rephrasings is not effective in current settings and therefore do not include this in the final model.

\vspace*{-2mm}

\section{Question Generation}
\label{ss:question_generation}
\begin{figure*}[htbp]
\begin{center}
\includegraphics[width=1.0\textwidth]{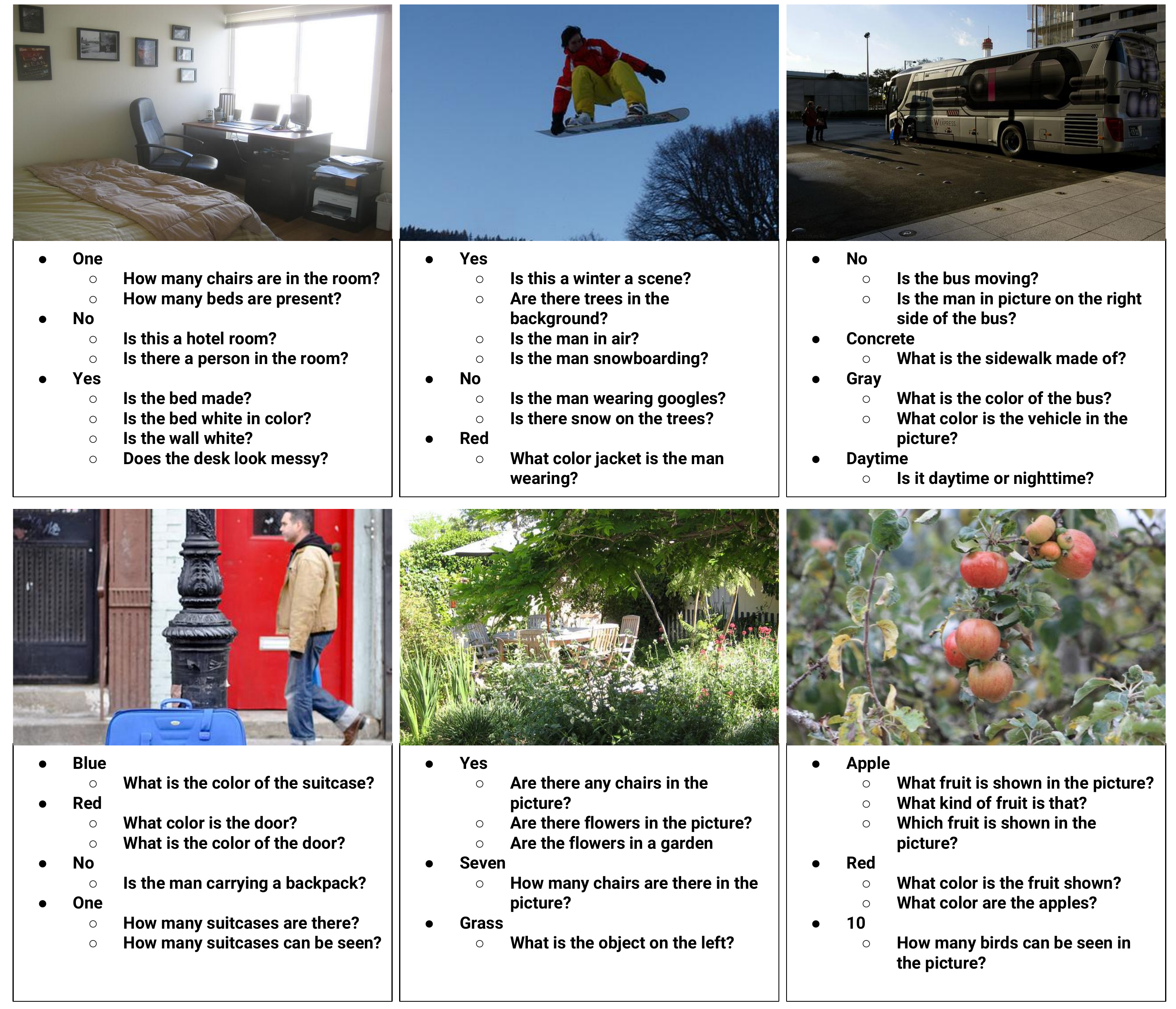}
\end{center}
   \caption{Qualitative examples of answer conditioned question generation by our VQG module.}
\label{fig:generated}
\end{figure*}

    Fig~\ref{fig:generated} shows qualitative examples of answer conditioned questions generated by our VQG model.
    Our VQG model is able to correctly generate answer conditioned questions for a wide range of answers ranging from numbers, to colors and even yes/no.
\vspace*{-2mm}

\section{Hyperparameters}
\label{ss:hyperparams}

We use the default hyperparameters as described in publicly available implementations of MUTAN~\cite{ben2017mutan}, BUTD~\cite{anderson2018bottom}, Pythia~\cite{pythia18arxiv} and BAN~\cite{Kim2018}.
When using these models as base VQA models to train cycle consistent variants of them, we use the same parameters for the VQA model. For the the VQG model we use $T_{sim}{=}0.9$, $\lambda_{G}{=}1.0$, $\lambda_{C}{=}0.5$ and $A_{iter}{=}5500$.
While some models use adaptive learning rates for their base VQA models, the VQG model is always trained with a fixed learning rate of $0.0005$. In case of BAN and Pythia, we also clip the gradients whose $L_{2}$ norm is greater than $0.25$. 

\end{document}